\documentclass[runningheads]{llncs}

\usepackage[english]{babel}
\usepackage{amsmath}
\usepackage{amsfonts}
\usepackage{amssymb}
\usepackage{graphicx}
\usepackage[colorlinks = true,
            linkcolor = black,
            urlcolor  = blue,
            ]{hyperref}
\usepackage{authblk}
\usepackage{subcaption}
\usepackage{booktabs}
\usepackage{array}
\usepackage[font={small,it}]{caption}
\usepackage{nameref}
\usepackage{mathpartir}
\usepackage{multicol}
\usepackage{stmaryrd}
\usepackage{enumitem}
\usepackage[nounderscore]{syntax}
\usepackage{fontawesome5}

\usepackage[T1]{fontenc}
\usepackage{lmodern}

\usepackage{tikz}
\usetikzlibrary{
    arrows.meta,      % For arrow styles
    positioning,      % For relative positioning of nodes
    shapes.geometric, % For shapes like rounded rectangles
    trees             % For drawing tree structures
}

\usepackage{xcolor}
\definecolor{lightgreen}{rgb}{0.8,1.0,0.8}
\definecolor{red}{rgb}{0.8,0,0}
\definecolor{green}{rgb}{0,0.8,0}
\definecolor{blue}{rgb}{0,0,1}
\definecolor{codegreen}{rgb}{0,0.6,0}
\definecolor{codegray}{rgb}{0.5,0.5,0.5}
\definecolor{codepurple}{rgb}{0.58,0,0.82}
\definecolor{backcolour}{rgb}{0.95,0.95,0.92}
\definecolor{keywordblue}{rgb}{0.13,0.13,1}
\definecolor{stringred}{rgb}{0.8,0,0}

\usepackage{listings}

\lstdefinestyle{lbnf}{
    backgroundcolor=\color{backcolour},   
    commentstyle=\color{codegreen},
    keywordstyle=\color{keywordblue}\bfseries,
    numberstyle=\tiny\color{codegray},
    stringstyle=\color{stringred},
    basicstyle=\ttfamily\footnotesize,
    breakatwhitespace=false,         
    breaklines=true,                 
    captionpos=b,                    
    keepspaces=true,                 
    numbers=left,                    
    numbersep=5pt,                  
    showspaces=false,                
    showstringspaces=false,
    showtabs=false,                  
    tabsize=1,
    frame=lines,
    xleftmargin=2em,
    framexleftmargin=1.5em,
    escapeinside={\%*}{*}
}

\lstdefinestyle{bash}{
    backgroundcolor=\color{backcolour},   
    commentstyle=\color{codegreen},
    keywordstyle=\color{keywordblue}\bfseries,
    numberstyle=\tiny\color{codegray},
    stringstyle=\color{stringred},
    basicstyle=\ttfamily\footnotesize,
    breakatwhitespace=false,         
    breaklines=true,                 
    captionpos=b,                    
    keepspaces=true,                 
    numbers=left,                    
    numbersep=5pt,                  
    showspaces=false,                
    showstringspaces=false,
    showtabs=false,                  
    tabsize=2,
    frame=lines,
    xleftmargin=2em,
    framexleftmargin=1.5em,
    escapeinside={\%*}{*} 
}

\lstnewenvironment{code}[1][]%
{
    \noindent
    \minipage{\linewidth} 
    \vspace{0.5\baselineskip}
    \lstset{basicstyle=\ttfamily\footnotesize,frame=single,#1}
}
{
    \endminipage
}

\newcommand{\inlinevnn}[1]{\texttt{#1}}

\usepackage[textsize=tiny]{todonotes}

\usepackage{todonotes}

\newcommand{\vnnlib}{VNN-LIB}

\newcommand{\nilList}{[\:]}
\newcommand{\consList}[2]{#1 :: #2}

\newcommand{\mgrammar}[1]{\text{\small \textit{#1}}}
\newcommand{\set}{\text{Set}}
\newcommand{\field}[2]{#1.\text{#2}}

\newcommand{\sem}[1]{\llbracket #1 \rrbracket}

\newcommand{\real}{\ensuremath{\mathbb{R}}}

\newcommand{\networkTheoryVar}{\Psi}
\newcommand{\missing}{\ensuremath{\square}}

\newcommand{\elementVar}{d}
\newcommand{\modelVar}{m}
\newcommand{\tensorVar}{t}

\newcommand{\shapeVar}{s}
\newcommand{\elementTypeVar}{\tau}
\newcommand{\tensorTypeVar}{\delta}
\newcommand{\modelTypeVar}{\gamma}

\newcommand{\outputNodesFn}{\text{networkOutputs}}

\newcommand{\semApp}[2]{\sem{#2}_\text{#1}}
\newcommand{\semElementType}[1]{\semApp{elementType}{#1}}
\newcommand{\semTensorType}[1]{\semApp{tensorType}{#1}}
\newcommand{\semTensor}[1]{\semApp{tensor}{#1}}

\newcommand{\semModel}[1]{\semApp{model}{#1}}

\newcommand{\semElementTypeAbs}{\semApp{elementType}{\cdot}}
\newcommand{\semTensorAbs}{\semApp{tensor}{\cdot}}
\newcommand{\semModelAbs}{\semApp{model}{\cdot}}

\newcommand{\networkTheoryVarParam}{^\networkTheoryVar}

\newcommand{\semTensorTheory}[1]{\semTensor{#1}\networkTheoryVarParam}

\newcommand{\semModelTheory}[1]{\semModel{#1}\networkTheoryVarParam}

\newcommand{\queryVar}{q}
\newcommand{\versionVar}{v}
\newcommand{\nameVar}{v}
\newcommand{\equivVar}{e}
\newcommand{\onnxNameVar}{u}

\newcommand{\networkDeclVar}{n}
\newcommand{\networkDeclSet}{N}

\newcommand{\inputDeclVar}{i}
\newcommand{\inputDeclSet}{I}

\newcommand{\hiddenDeclSet}{H}

\newcommand{\outputDeclSet}{O}

\newcommand{\assertionVar}{a}
\newcommand{\assertionSet}{A}

\newcommand{\namedNetworkExpansion}[1]
{
\texttt{declare-network} ~
#1 ~
\equivVar ~
\inputDeclSet ~
\hiddenDeclSet ~
\outputDeclSet
}

\newcommand{\networkExpansion}{\namedNetworkExpansion{\nameVar}}

\newcommand{\queryExpansion}
{
\versionVar ~
\networkDeclSet ~
\assertionSet
}

\newcommand{\inputExpansion}
{
\texttt{declare-input} ~
\nameVar ~
\elementTypeVar ~
\shapeVar
}

\newcommand{\hiddenExpansion}
{
\texttt{declare-hidden} ~
\nameVar ~
\elementTypeVar ~
\shapeVar ~
\onnxNameVar
}

\newcommand{\outputExpansion}
{
\texttt{declare-output} ~
\nameVar ~
\elementTypeVar ~
\shapeVar
}
\newcommand{\assertCtx}{\Gamma}

\newcommand{\assertEnv}{\Delta}
\newcommand{\semQuery}[1]{\semApp{query}{#1}}

\newcommand{\semArith}[1]{\semApp{arith}{#1}^\assertEnv}
\newcommand{\semBool}[1]{\semApp{bool}{#1}^\assertEnv}
\newcommand{\semAssert}[1]{\semApp{assert}{#1}^\assertEnv}

\newcommand{\networkDeclType}{\text{modelType}}

\newcommand{\networkImplementation}{\modelVar}
\newcommand{\networkImplementationSet}{M}

\newcommand{\networkInput}{x}
\newcommand{\networkInputSet}{X}

\usepackage{listings}

\begin{document}
\title{VNN-LIB 2.0: Rigorous Foundations for Neural Network Verification}

% Qui dentro metti il blocco reale degli autori
\author{
Ann Roy\inst{1}
\and
Allen Antony\inst{1}
\and
Andrea Gimelli\inst{2}\orcidID{0009-0008-1372-7694}
\and
Matthew L. Daggitt\inst{1}\orcidID{0000-0002-2552-3671} \faIcon{envelope}
}

\institute{
University of Western Australia, Perth, Australia\\
\email{matthewdaggitt@gmail.com} \faIcon{envelope}
\and
University of Genoa, Genoa, Italy
}
\authorrunning{Roy et al.}

\maketitle              
%

%\vspace{-3em}

\begin{abstract}
Neural network verification is an active and rapidly maturing research area, with a growing ecosystem of solvers and tools. The VNN-LIB standard was introduced to support interoperability in this ecosystem, but Version~1.0 has several serious short-comings as a formal foundation: it lacks a precise syntax, semantics, and type system, offers limited expressivity, and relies on externally defined ONNX models whose semantics are informal and constantly evolving. The latter distinguishes VNN-LIB from established standards such as SMT-LIB, where queries are self-contained and have fixed semantics.

In this paper we address these challenges by developing the theoretical foundations of VNN-LIB~2.0. Our key contribution is the introduction of the notion of a \emph{network theory}, which abstractly characterises the minimal semantic interface required from a neural network model format. This abstraction enables VNN-LIB to be defined independently of any specific ONNX version while remaining compatible with evolving model representations. Building on this foundation, we present a formal syntax for a more expressive query language, a type system for it over the numeric domains provided by the network theory, and finally a formal semantics. To ensure internal consistency, the standard is mechanised in the Agda theorem prover. VNN-LIB~2.0 therefore provides robust and rigorous foundations for trustworthy neural network verification.
\end{abstract}
\section{Introduction}
\label{sec:introduction}

\begin{figure}[t]
    \begin{minipage}[c]{0.72\textwidth}
        \begin{lstlisting}[style=lbnf]
; Input bounds
(assert (and (<= 0.0 X0) (<= X0 0.0))
(assert (and (<= 0.0 X1) (<= X1 0.0))
...
(assert (and (<= 0.0 X9) (<= X9 0.0))

; Output bounds
(assert (<= Y0 Y1))
\end{lstlisting}
    \end{minipage}%
    \begin{minipage}[c]{0.25\textwidth}
        \centering
        \includegraphics[height=3.5cm,alt={A visualisation of a neural network model with a `MatMul` and an `Add` node.}]{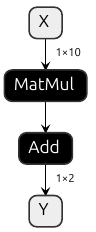}
    \end{minipage}
    \caption{A simple \vnnlib{} 1.0 query and accompanying neural network model. In \vnnlib{}~1.0, it is assumed that variables beginning with \texttt{X} represent elements of the input tensor, and variables beginning with \texttt{Y} represent elements of the output tensor.}
    \label{fig:vnnlib-1.0-query}
\end{figure}

As the field of neural network verification matures, the need for robust interoperability between solvers~\cite{ivanov2019verisig,lopez2023nnv,muller2022prima,NCubeV,wang2021beta,wu2024marabou} and higher-level tools~\cite{daggitt2025vehicle,girardsatabin2022caisar,shriver2021dnnv} is increasingly urgent.
In the SMT community, the SMT-LIB query language~\cite{barrett2010smt} has played a central role in enabling such interoperability by providing a solver-independent, formally specified interface between SMT solvers and a rich ecosystem of front-end tools.
Analogously, the VNN-LIB standard aims to serve as a common query language for neural network verification, allowing higher-level tools to target a single specification format.
VNN-LIB~1.0~\cite{demarchi2023supporting} is now widely adopted, and many solvers use it to compete annually in the VNN-COMP competition~\cite{brix2023first,kaulen20256th}. 
An example VNN-LIB 1.0 query is shown in Figure~\ref{fig:vnnlib-1.0-query}.

However, as outlined in recent papers~\cite{cordeiro2025neural,johnson2025neural}, Version~1.0 of VNN-LIB exhibits serious shortcomings as a rigorous foundation for neural network verification:

\begin{enumerate}
    \item \label{problem:syntax}
    \textbf{Lack of formal syntax}.  
    Version 1.0 does not formally describe the syntax of the VNN-LIB query language. As a result, solvers have implemented mutually incompatible variants, undermining the goal of interoperability.

    \item \label{problem:expressivity}
    \textbf{Lack of expressivity}.  
    Version 1.0 assumes a single neural network with a single input and single output. In practice, many verification tasks involve multiple networks~\cite{paulsen2020reludiff} or multiple applications of the same network~\cite{athavale2024verifying}. Until now, the community has worked around this limitation by merging ONNX models prior to verification. However, this approach is unsatisfactory as it prevents solvers from sharing bounds across copies of the same network and means that the ONNX artifact being verified is not the same as the set of ONNX artifacts being deployed and executed at runtime.

    \item \label{problem:types}
    \textbf{Lack of a type system}.
    Numeric soundness is a well-known challenge in neural network verification~\cite{jia2021exploiting}. While neural networks are typically executed using finite-precision floating-point arithmetic, they are often treated as real-valued by solvers during verification. Version 1.0 provides no mechanism for the user to specify what numeric types they expect the solver to use, providing ample opportunities for ambiguity and potential unsoundness.
    
    \item \label{problem:semantics}
    \textbf{Lack of semantics}.
    Version 1.0 does not define a formal semantics for the query language, i.e. an algorithm for deriving the precise mathematical problem represented by a given query. Consequently, it is not possible to formally prove the correctness of (i)~solvers that consume VNN-LIB, (ii)~higher-level tools that generate VNN-LIB, or (iii)~optimisations and transformations of VNN-LIB queries. It also hinders the development of proof certificates~\cite{elboher2025abstraction}.
\end{enumerate}
Given these issues, if the community is to build robust and interoperable neural network verification tools that users can have confidence in, there is a clear need for a more expressive and theoretically rigorous standard.

\subsection{Challenges specific to VNN-LIB}

At first glance, Problems~\ref{problem:syntax}--\ref{problem:semantics} resemble issues that have been successfully addressed by the SMT-LIB standard~\cite{barrett2010smt}.
However, the crucial difference is that SMT-LIB queries are self-contained descriptions of the problem being solved, whereas VNN-LIB queries specify properties of neural network models that are external to the query.
This separation is deliberate, as it enables verification to be performed directly on the executable model artifacts used in deployment, rather than on a non-executable re-encoded representation embedded in the query. However, it substantially complicates the resolution of Problems~\ref{problem:syntax}--\ref{problem:semantics}.

The neural network verification community has adopted ONNX~\cite{onnxruntime} as the standard format for such models, and the syntax and semantics of VNN-LIB therefore necessarily depend on those of ONNX.
Addressing Problems~\ref{problem:syntax}--\ref{problem:semantics} requires making this dependency explicit and precise. 
Problem~\ref{problem:semantics} poses a particular challenge, as ONNX itself currently lacks a formal semantics\footnote{In 2024, the ONNX Safety-Related
Profile working group was formed to develop the semantics of a subset of ONNX. As of 2026, no release date has been announced.}. 
Moreover, even if a formal semantics for ONNX were available, the ONNX standard will continue to evolve. Consequently, the syntax and semantics of \vnnlib{} must be defined in a way that accommodates the evolution of ONNX, without requiring a new version of \vnnlib{} for each ONNX release.

\subsection{Our contributions}

In this paper, we present the theoretical foundations of Version~2.0 of the VNN-LIB standard that allow us to precisely define the syntax, typing and semantics of the VNN-LIB query language, while overcoming the challenges identified above.

In Section~\ref{sec:network-theory}, we introduce the novel notion of a network theory, $\networkTheoryVar$: an abstract specification capturing the minimal syntax and semantics required of a neural network model format in order to define VNN-LIB.
This abstraction serves as an interface against which VNN-LIB is defined, without committing to the concrete syntax or semantics of ONNX. 
It therefore decouples the VNN-LIB standard from the evolution of ONNX, allowing a single version of VNN-LIB to be instantiated with multiple network theories, each corresponding to a different version of ONNX. 
As a result, the absence of a formal semantics for ONNX does not prevent us from defining a rigorous semantics for VNN-LIB.
    
In Section~\ref{sec:syntax}, we present a formal grammar for the VNN-LIB 2.0 query language that increases the expressivity of the standard by supporting (i)~multiple networks and network applications, (ii)~networks with multiple inputs/outputs, (iii)~access to hidden layers, and (iv)~explicit type annotations for networks. 
These extensions address Problems~\ref{problem:syntax} and~\ref{problem:expressivity}.
In Section~\ref{sec:typing}, we define a type system for the query language that enforces numeric type soundness with respect to the numeric types provided by the network theory~$\networkTheoryVar$, thereby addressing Problem~\ref{problem:types}.
In Section~\ref{sec:semantics}, we give a formal semantics for the VNN-LIB query language parameterised by the semantics supplied by $\networkTheoryVar$, solving Problem~4. 
In Section~\ref{sec:real-queries}, we demonstrate the expressive power of the network theory abstraction by showing how a simple transformation of $\networkTheoryVar$ yields a corresponding instance of the syntax, typing, and semantics for VNN-LIB queries with real-valued semantics.
    
Finally, to validate the internal consistency of the above, we mechanise the complete standard in the Agda interactive theorem prover. The resulting library, provided as an accompanying artifact, constitutes a canonical reference implementation of VNN-LIB~2.0 and can be used as a foundation for machine-checked proofs of correctness for solvers, query optimisations, and higher-level tools.

Together, these contributions ensure VNN-LIB~2.0 is a precise, extensible standard that provides a rigorous theoretical foundation upon which the neural network verification community can build interoperable and trustworthy tools. 
\section{Network theories}
\label{sec:network-theory}

\begin{figure}[p]
	\newcommand{\grammarShrink}{\vspace{0em}}
	
	\begin{subfigure}{\textwidth}
		\input{diagrams/network-theory-types.tex}
		\vspace{-0.7em}
		\caption{Abstract grammar for neural network model types.}
		\label{fig:onnx-type-syntax}
	\end{subfigure}
	\\
	\vspace{2.3em}
	\\
	\begin{subfigure}{\textwidth}
		\input{diagrams/network-theory-expr.tex}
		\vspace{0.3em}
		\caption{Abstract grammar and functions for neural network models.}
		\label{fig:onnx-expr-syntax}
	\end{subfigure}
	\\
	\vspace{1em}
    \\
	\begin{subfigure}{\textwidth}
		\input{diagrams/network-theory-typing-rules.tex}
		\caption{Abstract type system for network models.}
		\label{fig:onnx-types}
	\end{subfigure}
	\\
	\vspace{1em}
    \\
	\begin{subfigure}{\textwidth}
		\input{diagrams/network-theory-semantics.tex}
		\caption{Abstract semantics for network models.}
		\label{fig:onnx-semantics}
	\end{subfigure}
	\caption{The definition of a network theory, i.e. the minimal signature for an abstract implementation of ONNX that allows the syntax and semantics of VNN-LIB to be defined. The \missing{} symbol indicates what needs to be defined by the ONNX standard to instantiate the theory. Superscript $^+$ indicates a list of one or more.}
	\label{fig:onnx-signature}
\end{figure}

We begin by defining the concept of a \emph{network theory} -- the minimal set of syntax, typing judgments and semantics for the neural network model format necessary to define the syntax and semantics of VNN-LIB~2.0. 
Figures~\ref{fig:onnx-type-syntax}~\&~\ref{fig:onnx-expr-syntax} describe syntax of the model format and its type system. A concrete implementation of a network theory, $\networkTheoryVar$, (e.g. for ONNX  v1.20.0) is required to define:
\begin{enumerate}
\item $\mgrammar{<elementType>}$ - a set of numeric element types supported by the model format (i.e.~ONNX element types \texttt{float64}, \texttt{int32}, etc.).
\item $\mgrammar{<tensor>}$ - the representation of tensors (i.e. ONNX \texttt{TensorProto} objects).
\item $\mgrammar{<model>}$ - the representation of neural network models (i.e. the ONNX \texttt{ModelProto} objects serialised to disk as \texttt{.onnx} files).
\item $\mgrammar{<nodeOutput>}$ - a method of referencing node outputs (i.e. according to the ONNX standard, any string compliant with the C90 identifier syntax rules).
\item $\outputNodesFn$ - a function that maps a model to a list of references to its final outputs (i.e. in ONNX the accessor \texttt{model.graph.output}).
\end{enumerate}
Figure~\ref{fig:onnx-types} describes typing judgments over the neural network model format provided in Figure~\ref{fig:onnx-expr-syntax}. A concrete implementation $\networkTheoryVar$ is required to define:
\begin{enumerate}
\item \textsc{(Element)} - a judgement that a numeric string~$\elementVar$ is a valid element of a provided element type~$\elementTypeVar$. For example, if the network theory provided types \texttt{float64} and \texttt{int32}, it would be expected (although not required) that the number~`1' could be judged as of type \texttt{float64} and \texttt{int32}, and `1.1' could be judged as of type \texttt{float64} but not \texttt{int32}.
\item \textsc{(Tensor)} - a judgement that a tensor~$\tensorVar$ is of type~$\tensorTypeVar$, where $\tensorTypeVar$ is a description of the element type and shape of the tensor.
\item \textsc{(Model)} - a judgment that a model~$\modelVar$ is of type~$\modelTypeVar$, where $\modelTypeVar$ is a description of the types of the tensors it accepts as input and produces as output.
\item \textsc{(NodeOutput)} - a judgment that a model~$\modelVar$ contains a node that produces some tensor called~$\onnxNameVar$ of type~$\tensorTypeVar$ as output.
\item \textsc{(Isomorphic)} - a judgment that two models are isomorphic.
\end{enumerate}
Finally, Figure~\ref{fig:onnx-semantics} describes the semantics over the well-typed neural network models. A concrete implementation $\networkTheoryVar$ is required to define:
\begin{enumerate}
\item $\semElementTypeAbs$ - a function that maps each element type to the mathematical set of values it represents (e.g. it would be expected that the ONNX type \texttt{float32} would be mapped to the set of all 32-bit floating point numbers).
\item $\semTensorAbs$ - a function mapping a syntactic tensor $\tensorVar$ of type $\tensorTypeVar$ to the mathematical tensor that it represents.
\item $\semModelAbs$ - a function that takes a model, $\modelVar$, and a reference to an output of one of the model's nodes, $\onnxNameVar$, as input and returns another function that represents the semantics of the model for that output, i.e. the precise mathematical function that $m$ uses to compute the value of $u$.
\item $\sem{\leq}$, ..., $\sem{\times}$ - the semantics of basic pointwise comparison and arithmetic operations, defined for tensors of any shape and numeric type provided by~$\networkTheoryVar$.
\end{enumerate}
Throughout the remainder of this paper, we define the syntax, typing and semantics of the \vnnlib{}~2.0 query language relative to an abstract network theory~$\networkTheoryVar$.
Instantiating~$\networkTheoryVar$ with a concrete description of a particular version of ONNX then yields a complete definition of the query language.
Consequently, if a future ONNX release introduces changes (e.g. a new floating-point type), only the instantiation of~$\networkTheoryVar$ needs to be updated, rather than the \vnnlib{} standard itself.
Another consequence of this parameterisation is that it emphasises the need to record both the ONNX version and the \vnnlib{} version used in any verification task -- information that is sometimes omitted in current practice.

In practice, each concrete implementation of $\networkTheoryVar$ corresponds to a ONNX opset version. 
It is a deliberate design choice that the \vnnlib{} query file does not explicitly reference the ONNX opset version. 
Instead, the opset version, and hence the network theory to be used, should be extracted from the accompanying ONNX model file(s) passed to the solver alongside the query. 
Currently if multiple model files using different ONNX opsets are passed to the solver then the solver should error\footnote{There are ongoing discussions on how to best to loosen this restriction so as to support backwards compatible opsets in future versions of the standard.}. 

Note that nothing in the definition of a network theory is inherently tied to ONNX. 
It would be equally valid to use $\networkTheoryVar$ to interpret \vnnlib{} with respect to other model formats, such as PyTorch's \texttt{.pt}/\texttt{.pth} format~\cite{Ansel_PyTorch_2_Faster_2024}. Nevertheless, given the widespread adoption of ONNX within the verification community, \vnnlib{}~2.0 continues to assume ONNX will be used to instantiate $\networkTheoryVar$ in practice.
\section{Syntax}
\label{sec:syntax}

\begin{figure}[t]
	\setlength{\grammarindent}{6.5em}
	\input{diagrams/vnnlib-syntax.tex}
    \caption{The syntax of \vnnlib{} queries. The superscript $^*$~means zero or more, $^+$~means one or more, $^?$~means optional and $\networkTheoryVarParam$~means the piece of syntax is defined by the underlying network theory $\networkTheoryVar$ as described in Figures~\ref{fig:onnx-type-syntax}~\&~\ref{fig:onnx-expr-syntax}.}
    \label{fig:vnnlib-syntax}
\end{figure}

\begin{figure}[t]
    \begin{minipage}[c]{0.72\textwidth}
        \begin{lstlisting}[style=lbnf]
(vnnlib-version <2.0>)

(declare-network myNetwork
  (declare-input  X float32 [1,10])
  (declare-output Y float32 [1,2]))

(assert (>= X[0,2] 0.0))
(assert (<= X[0,2] 1.0))
(assert (<= Y[0,1] 0.5))\end{lstlisting}
    \end{minipage}%
    \begin{minipage}[c]{0.25\textwidth}
        \centering
        \includegraphics[height=4cm,alt={A visualisation of a neural network model with a `MatMul` and an `Add` node with an input and an output labelled `X` and `Y` respectively.}]{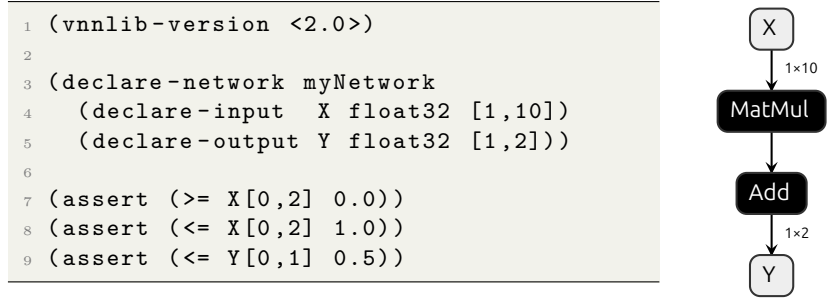}
    \end{minipage}
    \caption{A simple \vnnlib{} query which declares a network with a single input and output tensor. An example of one of the many possible ONNX models compatible with this declaration is shown on the right. Note that the variable names \texttt{X} and \texttt{Y} for the declared inputs and outputs do not have to match the node names in the ONNX model.}
    \label{fig:simple-query}
\end{figure}

Figure~\ref{fig:vnnlib-syntax} shows the syntax of the \vnnlib{}~2.0 query language. Whereas \vnnlib{}~1.0 queries only contained assertions, \vnnlib{}~2.0 queries have three parts: 
\begin{enumerate}
\item \textbf{Version} - the VNN-LIB version the query was written with, which allows tooling to detect incompatibilities and provide better error messages.
\item \textbf{Networks} - a non-empty list of network declarations that allows the introduction of variables representing inputs and outputs of the networks.
\item \textbf{Assertions} - a list of assertions that reference and constrain the variables introduced by the network declarations.
\end{enumerate}
For simplicity, \vnnlib{} 2.0 does not currently permit the interleaving of network declarations and assertions. Figure~\ref{fig:simple-query} shows an example of a simple query.

\subsection{Network Declarations}
\label{sec:network-declarations}

A network is introduced by the keyword \texttt{declare-network}, followed by a user-defined name for the network. The network name is used to explicitly associate an ONNX model file with the network declaration when calling the solver. 
Next is a list of declarations for the network's inputs and outputs.
An input is declared using \texttt{declare-input}, followed by a variable name, an element type from the network theory $\networkTheoryVar$ (e.g., \texttt{float64}, \texttt{int32}), 
and the shape of the tensor. Similarly, an output variable is declared using \texttt{declare-output}. 

It is a deliberate design choice that a network declaration acts only as an interface and does not refer to a specific ONNX model. 
In particular, the declared network name need not match the name of the ONNX model, the names of the declared input and output variables need not match the names of the corresponding ONNX nodes, and the declaration imposes no constraints on the internal architecture of the model.
From a theoretical perspective, this reflects the principle that a VNN-LIB query should constrain the mathematical function represented by a network rather than the internal details of a particular model format. 
From a practical perspective, this avoids the need to rewrite queries when a model is refactored or re-exported, or when changes in the toolchain alter the structure or naming of the ONNX file. In turn, this improves the robustness of benchmarks and of the VNN-COMP competition infrastructure.

\subsection{Assertions}

Variables introduced by the network declarations are constrained by assertions. Assertions are quantifier-free logical formulas expressed in standard SMT-LIB-like syntax where expressions and operations are composed using bracketed reverse Polish notation. \vnnlib{}~2.0 provides the same logical connectives, relational comparisons, and arithmetic operations as \vnnlib{}~1.0.

Unlike \vnnlib{} 1.0, standard indexing notation is used to refer to a specific element of a tensor variable. For example, \inlinevnn{X[0,2]} in Figure~\ref{fig:simple-query} refers to the value of the element of the input tensor at row~0, column~2. Indices are zero based and the number of indices provided must be equal to the number of dimensions of the variable, i.e. partial indexing is not allowed.

\subsection{Complex Network Declarations}
\label{sec:complex-networks-decls}

Whereas \vnnlib{} 1.0 only allowed users to constrain the single input and single output of a single network, \vnnlib{} 2.0 supports more complex queries.

\paragraph{Multiple input and outputs.}

\begin{figure}[t]
    \centering
    \begin{lstlisting}[style=lbnf]
(declare-network multi_io_net
    (declare-input  image    float32 [1,3,224,224])
    (declare-input  metadata float32 [1,10])
    (declare-output bbox     int16   [1 4])
    (declare-output logits   float32 [1,1000]))\end{lstlisting}

    \vspace{0.5cm}
    \includegraphics[width=0.99\textwidth, alt={A visualisation of a neural network model with two inputs labelled `image` and `metadata` and two outputs labelled `bbox` and `logits`.}]{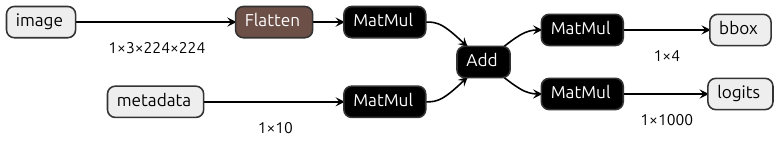}
    \caption{A \vnnlib{} network declaration with multiple inputs/outputs.}
    \label{fig:multi-inputs-outputs}
\end{figure}

Neural network that accept multi-modal input or produce multi-modal output are increasingly common~\cite{wang2021survey}. 
For example, Figure~\ref{fig:multi-inputs-outputs} shows a simple model that takes both an image and meta-data about that image as inputs, and outputs both a bounding box around an identified object of interest and a probability distribution over the possible class that the object belongs to. 
\vnnlib{}~2.0 supports queries over such networks by allowing a network declaration to declare an arbitrary number of input and output declarations. 
The declared inputs and outputs are mapped to the ONNX graph's inputs and outputs by matching the order of the declarations in the query to the order of nodes in the ONNX model file.

\paragraph{Hidden nodes.}
\label{sec:hidden-output-declarations}

\begin{figure}[h!]
    \begin{minipage}[c]{0.80\textwidth}
        \begin{lstlisting}[style=lbnf]   
(declare-network f
  (declare-input  X float32 [1,28,28])
  (declare-hidden Z float32 [1,128] "hidden")
  (declare-output Y float32 [1,10]))
  
(assert (>= X[0,0,0] 0.0))
(assert (>= Z[0,0]   0.5))
(assert (>= Y[0,0]   1.0))
\end{lstlisting}
    \end{minipage}%
    \begin{minipage}[c]{0.19\textwidth}
        \centering
        \includegraphics[height=4cm,alt={A visualisation of a neural network model with three nodes, with the one of the edges in the graph lablled as 'hidden'.}]{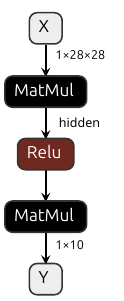}
    \end{minipage}
    \caption{A \vnnlib{} network declaration that declares a hidden node}
    \label{fig:hidden-node}
\end{figure}

In some cases it is desirable to constrain the result of intermediate computation at the output of hidden nodes within the network. For example, when reasoning about the encodings in an encoder-decoder architecture or when reasoning about attention mechanisms. This can be achieved by using the \texttt{declare-hidden} declaration, as shown in Figure~\ref{fig:hidden-node}. This declaration includes a variable name for use within the \vnnlib{} specification, its element type, its tensor shape, and lastly, the name of the desired intermediate tensor in the ONNX model that the variable refers to. The latter, a member of the \emph{<nodeOutput>}$\networkTheoryVarParam$ class supplied by the underlying network theory $\networkTheoryVar$, acts as a unique reference to an output of some node in the ONNX graph. Note that the variable represents an output value of the node, rather than the node itself, as ONNX nodes may produce multiple outputs. Multiple hidden nodes can be declared within a single network declaration.

\paragraph{Multiple networks.}
\label{sec:multiple-network}

\begin{figure}[ht]
    \begin{minipage}[c]{0.67\textwidth}
        \begin{lstlisting}[style=lbnf]
(declare-network teacher
  (declare-input  TX float32 [1,32])
  (declare-output TY float32 [1,2]))

(declare-network student
  (declare-input  SX float16 [1,32])
  (declare-output SY float16 [1,2]))\end{lstlisting}
    \end{minipage}
    \begin{minipage}[c]{0.32\textwidth}
        \centering
        \includegraphics[height=4cm, alt={A visualisation of a neural network model with two nodes representing the student network.}]{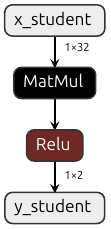}
        \includegraphics[height=4cm, alt={A visualisation of a more complex neural network model with three nodes representing the teacher network.}]{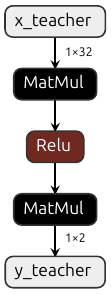}
    \end{minipage}
    \caption{\vnnlib{} network declarations allowing the referencing of multiple networks.}
    \label{fig:multiple-networks}
\end{figure}

\begin{figure}[ht]
    \begin{minipage}[c]{0.67\textwidth}
        \begin{lstlisting}[style=lbnf]
(declare-network f
  (declare-input  A float32 [1,10])
  (declare-output B float32 [1,2]))

(declare-network f-copy
  (equal-to f)
  (declare-input  C float32 [1,10])
  (declare-output D float32 [1,2]))\end{lstlisting}
    \end{minipage}
    \begin{minipage}[c]{0.32\textwidth}
        \centering
        \includegraphics[height=4cm, alt={A visualisation of a neural network model with two nodes.}]{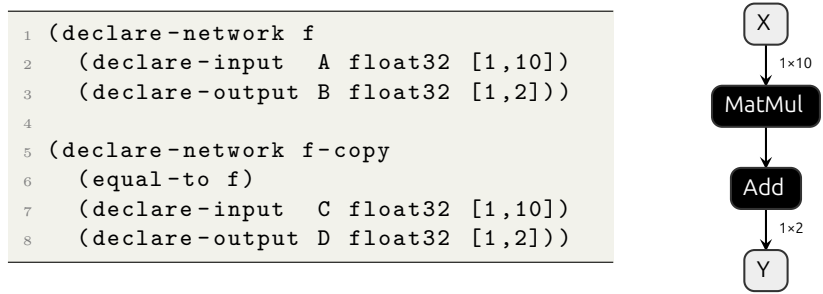}
    \end{minipage}
    \caption{A pair of \vnnlib{} network declarations that reference the same ONNX model.}
    \label{fig:multiple-equal-networks}
\end{figure}

\begin{figure}[ht]
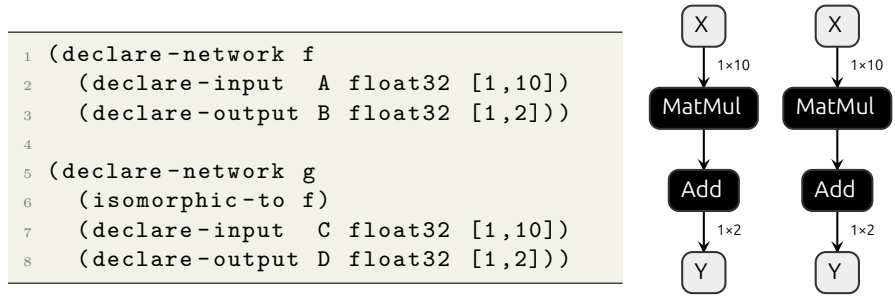

    \begin{minipage}[c]{0.68\textwidth}
        \begin{lstlisting}[style=lbnf]
(declare-network f
  (declare-input  A float32 [1,10])
  (declare-output B float32 [1,2]))

(declare-network g
  (isomorphic-to f)
  (declare-input  C float32 [1,10])
  (declare-output D float32 [1,2]))
\end{lstlisting}
    \end{minipage}
    \begin{minipage}[c]{0.31\textwidth}
        \centering
        \includegraphics[height=4cm, alt={A visualisation of a neural network model with two nodes.}]{imgs/simple_net.onnx.pdf}
        \vspace{0.5cm} 
        \includegraphics[height=4cm, alt={A visualisation of the same neural network model with two nodes.}]{imgs/simple_net.onnx.pdf}
    \end{minipage}
    \caption{A pair of \vnnlib{} network declarations referencing isomorphic ONNX models.}
    \label{fig:multiple-isomorphic-networks}
\end{figure}

In some scenarios it is desirable to constrain the interaction of multiple neural networks, e.g. teacher-student networks or encoder-decoder architectures. \vnnlib{}-2.0 supports this by allowing a query to contain multiple network declarations, as shown in Figure~\ref{fig:multiple-networks}.
Often multiple \texttt{declare-network} declaration will map to the same network model (e.g. proving monotonicity~\cite{liu2020certified} or global robustness~\cite{kabaha2024verification}). In this case, the \texttt{equal-to} declaration may be used as shown in Figure~\ref{fig:multiple-equal-networks}.  
In other cases, each network declaration will be mapped to a network with the same graph structure but with different weights (e.g. proving local equivalence after retraining or quantisation~\cite{henzinger2021scalable,huang2024towards}). In this case, the \texttt{isomorphic-to} declaration may be used as shown in Figure~\ref{fig:multiple-isomorphic-networks}. 
Providing this information allows the solver to share transfer its deductions (e.g. variable bounds) between the networks, thereby drastically improving efficiency.

\section{Typing}
\label{sec:typing}

We now define typing judgements for \vnnlib{} queries, i.e. a formal description of the constraints on the syntax described in Section~\ref{sec:syntax} that ensure a query is mathematically well-formed. 
Typing judgements are defined relative to a \emph{typing context}, $\assertCtx$, which maps the names of networks in scope to the corresponding network declarations.
A query $\queryVar = \queryExpansion$ is typed by first using the network declarations to construct $\assertCtx$ and then using $\assertCtx$ to type the assertions:
\begin{mathpar}
            \inferrule*[Right=Query]
            {
	            \emptyset \vdash \networkDeclSet : \assertCtx
	            \\
	            \assertCtx \vdash \assertionSet
            }
            {
                \vdash \queryExpansion
            }
\end{mathpar}\unskip

\subsection{Network Declarations}

Network declarations both extend the typing context and may refer to networks declared earlier in the context via \texttt{equal-to} and \texttt{isomorphic-to} statements.
We therefore define a judgement $\assertCtx_1 \vdash \networkDeclSet : \assertCtx_2$, which asserts that a list of network declarations~$\networkDeclSet$ is well-typed relative to an initial context~$\assertCtx_1$ and yields a resulting context~$\assertCtx_2$. This judgement is defined recursively as follows:
\begin{mathpar}
    \hspace{-6em}
    \inferrule*[Right=NetworkNil]
    {
    }
    {
        \assertCtx \vdash \nilList : \assertCtx
    }
    \hspace{6.5em}
    \inferrule*[Right=NetworkCons]
    {
        \assertCtx_1 \vdash \networkDeclVar
        \\
        \assertCtx_1[\field{\networkDeclVar}{name} \rightarrow \networkDeclVar] \vdash \networkDeclSet : \assertCtx_2
    }
    {
        \assertCtx_1
        \vdash \consList{\networkDeclVar}{\networkDeclSet} : 
        \assertCtx_2
    }
\end{mathpar}
\unskip
where $\nilList$ and $::$ represent the standard nil and cons for constructing lists.
In particular, an empty list of network declarations is always well-typed under any context and returns the context unaltered. A non-empty list of network declarations is well-typed if the head of the list can be typed under the current context, and the tail is well-typed under the current context extended by the head of the list.
The judgement that a network is well-typed is defined as follows:
\begin{mathpar}
    \hspace{-3em}
    \inferrule*[Right=Network]
    {
        \assertCtx, \inputDeclSet, \hiddenDeclSet, \outputDeclSet \vdash e
    }
    {
        \assertCtx \vdash \networkExpansion
    }
\end{mathpar}
\unskip
where the only requirement is that the equivalence statement $e$ is well-formed with respect to the context and input/hidden/output node declarations $I$/$H$/$O$. 

There are three possibilities for the equivalence statement. Firstly, if there is no equivalence statement then the statement is judged to be trivially well-formed. 
Next, if the network is marked as equal to some previous network in the context then both the element types and shapes of the input and output declarations for the two networks have to match:
\begin{mathpar}
    \inferrule*[Right=NEq, vskip=1ex]
    {
        \assertCtx[v] = \networkExpansion \qquad e = \varnothing
        \\
        \forall i . \: \field{\inputDeclSet_i}{shape} = \field{\inputDeclSet'_i}{shape} \wedge \field{\inputDeclSet_i}{type} = \field{\inputDeclSet'_i}{type}
        \\
        \forall i . \: \field{\outputDeclSet_i}{shape} = \field{\outputDeclSet'_i}{shape} \wedge \field{\outputDeclSet_i}{type} = \field{\outputDeclSet'_i}{type}
        \\
        \forall i, j . \: \field{\hiddenDeclSet_i}{name} = \field{\hiddenDeclSet'_j}{name} \Rightarrow \field{\hiddenDeclSet_i}{shape} = \field{\hiddenDeclSet'_j}{shape} \wedge \field{\hiddenDeclSet_i}{type} = \field{\hiddenDeclSet'_j}{type}
    }
    {
        \assertCtx, \inputDeclSet', \hiddenDeclSet', \outputDeclSet' \vdash \texttt{equal-to }\nameVar
    }
    \qquad
    \qquad
\end{mathpar}
\unskip
While equal networks do not necessarily need to declare the same set of hidden nodes, the element types and shape of any hidden node declarations that do share a common name must match. The last constraint is the network referred to by the \texttt{equal-to} statement must not contain an equivalence statement.

Finally, if the network is marked as isomorphic to some other network in the context then the shapes, but not the element types, of the input and output declarations for the two network have to match:
\begin{mathpar}
    \inferrule*[Right=NIso, vskip=1ex]
    {
        \assertCtx[v] = \networkExpansion \qquad e = \varnothing
        \\
        \forall i . \: \field{\inputDeclSet_i}{shape} = \field{\inputDeclSet'_i}{shape}
        \\
        \forall i . \: \field{\outputDeclSet_i}{shape} = \field{\outputDeclSet'_i}{shape}
    }
    {
        \assertCtx, \inputDeclSet', \hiddenDeclSet', \outputDeclSet' \vdash \texttt{equal-to }\nameVar
    }
\end{mathpar}
\unskip
There are no constraints on hidden node declarations as, although there must necessarily be a one-to-one mapping between the internal nodes of two isomorphic networks, in practice those isomorphic nodes may be assigned different names in the models.
As with equal networks, the network referred to by the \texttt{isomorphic-to} statement must not contain an equivalence statement.

The restrictions that an \texttt{equal-to} or \texttt{isomorphic-to} declaration cannot reference another network declaration that also contains an equivalence statement, was introduced make the dependencies easier to track by the solvers and immediately eliminate the possibility of cycles. 
It does not reduce the expressive power the language due to the transitivity of equality and isomorphism.

\subsection{Assertions}

Boolean expressions are type-checked relative to a typing context $\assertCtx$ as follows:
\begin{mathpar}
    \hspace{-6.5em}
    \inferrule*[Right=Assertion]
    {
        \assertCtx \vdash b
    }
    {
        \assertCtx \vdash \texttt{assert}~b
    }
    \hspace{5.8em}
    \inferrule*[Right=N-ary Boolean]{
        \assertCtx \vdash b_1 \quad ... \quad \assertCtx \vdash b_n \quad \lozenge \in \{\texttt{and}, \texttt{or}\}
    }{
        \assertCtx \vdash \lozenge~b_1~...~b_n
    }
    \\
    \hspace{-10em}
    \inferrule*[Right=Comparison]{
        \assertCtx \vdash a_1 : \tau \quad \assertCtx \vdash a_2 : \tau \quad \lozenge \in \{\texttt{>=}, \texttt{>}, \texttt{<}, \texttt{<=}, \texttt{==}, \texttt{!=}\}
    }{
        \assertCtx \vdash \lozenge~a_1~a_2
    }
\end{mathpar}
\unskip
The arithmetic operations typing judgements are defined as follows:
\begin{mathpar}
    \hspace{-2em}
    \inferrule*[Right=Negation]
    {
        \assertCtx \vdash a : \tau
    }
    {
        \assertCtx \vdash \texttt{-}~a : \tau
    }
    \hspace{7em}
    \inferrule*[Right=Arith]
    {
        \assertCtx \vdash a_1 : \tau \quad ... \quad \assertCtx \vdash a_n : \tau \quad \lozenge \in \{\texttt{+}, \texttt{-}, \texttt{×}\}
    }
    {
        \assertCtx \vdash \lozenge~a_1~...~a_n : \tau
    }
\end{mathpar}
\unskip
Some ONNX types can be soundly coerced between, e.g. a \inlinevnn{float16} can be added to a \inlinevnn{float32} if the expected result type is a \inlinevnn{float32}. However, the standard does not currently permit such coercions and arguments to arithmetic comparisons and operators must have the same type.
The typing judgements for constants defers to the typing judgement provided by the underlying network theory~$\networkTheoryVar$:
\begin{mathpar}
    \inferrule*[Right=Constant]
    {
        \vdash_\networkTheoryVar d : \tau
    }
    {
        \assertCtx \vdash d : \tau
    }
\end{mathpar}
\unskip
For example, one might expect that the constant \texttt{0.5} would be a member of the ONNX \texttt{float32} element type, but not of the ONNX \texttt{int32} element type.

Finally, the typing judgements for variables ensure they are of the correct type and the indices provided fit within the declared shape of the variable:
\begin{mathpar}
    \inferrule*[Right=Variable, vskip=1ex]
    {
        \exists \nameVar' . \; \assertCtx[\nameVar'] = \namedNetworkExpansion{\nameVar'}
        \qquad
        \forall i . \; l_i < \shapeVar_i
        \\
        \exists d \in \inputDeclSet \cup \hiddenDeclSet \cup \outputDeclSet . \;\;
        \field{d}{name} = v \wedge \field{d}{shape} = \shapeVar \wedge \field{d}{type} = \elementTypeVar  
    }
    {
        \assertCtx \vdash \nameVar~[l_1, ..., l_n] : \tau
    }
\end{mathpar}
\unskip
Note that variables therefore provide the source of typing for the whole arithmetic expression and one consequence of this is that comparisons that contain no variables (e.g. \texttt{(<= 0.0 1.0)}) are not typable.

\subsection{Models and Input Assignments}
\label{sec:model-typing}

At this point we have defined what it means for a query to be well-formed mathematically. However, in order to define the semantics in Section~\ref{sec:semantics}, we must also describe how the query relates to the inputs and outputs of the solver.
The input to a solver is a query~$\queryVar$ and a mapping
$\networkImplementationSet : \mgrammar{<name>} \rightarrow \mgrammar{<model>}\networkTheoryVarParam$
which associates the name of each network declaration in $\queryVar$ with a corresponding model from $\networkTheoryVar$. 
A mapping~$\networkImplementationSet$ is well-typed with respect to a query~$\queryVar = \queryExpansion$ if it can be judged to be well-formed with respect to the list of network declarations. 
\begin{equation*}
    \inferrule*[Right=ModelQuery]
    {
        \networkDeclSet
        \vdash 
        \networkImplementationSet 
    }
    {
        \queryExpansion
        \vdash 
        \networkImplementationSet
    }
\end{equation*}\unskip
A mapping is well-typed with respect to a list of network declarations if each network declaration is matched by a model:
\begin{equation*}
    \inferrule*[Right=ModelNil]
    {
    }
    {
        \nilList 
        \vdash 
        \networkImplementationSet 
    }
    \qquad
    \qquad
    \qquad
    \qquad
    \inferrule*[Right=ModelCons]
    {
        \networkImplementationSet,\networkDeclVar 
        \vdash \networkImplementationSet[\networkDeclVar.\text{name}]
        \\
        \networkDeclSet  
        \vdash 
        \networkImplementationSet 
    }
    {
        \consList{\networkDeclVar}{\networkDeclSet}  
        \vdash 
        \networkImplementationSet 
    }
    \qquad
    \qquad
    \quad
\end{equation*}
\unskip
A model $\networkImplementation$ is well-typed with respect to a network declaration $\networkDeclVar$ if (i) it is compatible with the equivalence statement, (ii) it can be typed according to the network declaration and (iii) it has hidden nodes of the correct type: 
\begin{equation*}
    \inferrule*[Right=ModelNetwork]
    {
        \networkImplementationSet,
        \equivVar
        \vdash 
        \networkImplementation
        \\
        \vdash_\networkTheoryVar \networkImplementation : \networkDeclType(\inputDeclSet, \outputDeclSet)
        \\
        \hiddenDeclSet 
        \vdash 
        \networkImplementation
    }
    {
        \networkImplementationSet,\; \networkExpansion 
         \vdash
         \networkImplementation
    }
    \qquad
    \qquad
    \qquad
\end{equation*}
\unskip
where the function $\networkDeclType : \mgrammar{<input>}^+ \times \mgrammar{<output>}^+ \rightarrow \mgrammar{<modelType>}\networkTheoryVarParam$ maps the input and output declarations to the type of the model in the obvious way.

A model $\networkImplementation$ is compatible with $\texttt{isomorphic-to}\,\nameVar$ and $\texttt{equal-to}\,\nameVar$ statements if it is related to the referenced network model $\nameVar$ in the appropriate way:
\begin{align*}
    \inferrule*[Right=ModelIso]
    {
        \networkImplementationSet[\nameVar] \cong_\networkTheoryVar \networkImplementation
    }
    {
        \networkImplementationSet, \;
        \texttt{isomorphic-to }\nameVar 
        \vdash 
        \networkImplementation
    }
    \qquad\qquad\qquad\quad
    \inferrule*[Right=ModelEqual]
    {
        \networkImplementationSet[\nameVar] = \networkImplementation
    }
    {
        \networkImplementationSet, \;
        \texttt{equal-to }\nameVar 
        \vdash 
        \networkImplementation  
    }
    \qquad
    \qquad
    \qquad
\end{align*}
\unskip
If no equivalence statement is provided, then any network implementation is compatible. Note that the equality judgment `$=$' in the $\texttt{equal-to }\nameVar$ case is \emph{not} parameterised by $\networkTheoryVar$ as it refers to absolute equality, i.e. the network models must actually be the \emph{same} model file, rather than merely being semantically equal. 

Finally a network model is compatible with a list of hidden node declarations if each declaration corresponds to an output of some node in the model of the correct type:
\begin{equation*}
    \inferrule*[Right=ModelHidden]
    {
        \networkImplementation \vdash_\networkTheoryVar \onnxNameVar : (\elementTypeVar, \shapeVar)
    }
    {
        \hiddenExpansion
        \vdash 
        \networkImplementation
    }
    \qquad
    \qquad
\end{equation*}
\unskip

As output, a solver either judges that the query $\queryVar$ is unsatisfiable or produces a satisfying assignment consisting of a mapping
$\networkInputSet : \mgrammar{<name>} \rightarrow \mgrammar{<tensor>}\networkTheoryVarParam$
which associates each input declaration in $\queryVar  = \queryExpansion$ with a corresponding tensor object from~$\networkTheoryVar$. 
A mapping~$\networkInputSet$ is well-formed with respect to a query~$\queryVar$ if it can be judged to be well-formed with respect to all the input declarations in $\queryVar$. 
\begin{equation*}
    \inferrule*[Right=InputAssignmentQuery]
    {
        \forall \networkDeclVar \in \networkDeclSet \; . \; \forall \inputDeclVar \in \field{\networkDeclVar}{inputs} \; . \; 
        \inputDeclVar
        \vdash 
        \networkInputSet 
    }
    {
        \queryExpansion
        \vdash 
        \networkInputSet
    }
    \qquad
    \qquad
    \qquad
    \qquad
\end{equation*}\unskip
Using the typing judgement \textsc{Tensor} from the underlying network theory in Figure~\ref{fig:onnx-signature}, the following typing judgment ensures that the input assignment~$\networkInputSet$ is well typed with respect to the network declarations in the query~$\queryVar$:
\begin{equation*}
    \hspace{-5em}
    \inferrule*[Right=InputAssignment]
    {
        {\begin{array}{c}
            \networkInputSet[\nameVar] \vdash_\networkTheoryVar (\elementTypeVar, \shapeVar)
        \end{array}}
    }
    {
        \inputExpansion
        \vdash 
        \networkInputSet 
    }
\end{equation*}\unskip

\section{Semantics}
\label{sec:semantics}

We now define the semantics of the \vnnlib{}~2.0 query language, i.e. a precise description of the mathematical satisfiability problem that a solver is attempting to answer when provided with a well-typed query $\queryVar$ and set of network models~$\networkImplementationSet$.

\paragraph{Environment.}
In order to determine whether a query $\queryVar$ over a set of models $\networkImplementationSet$ is satisfied by an input assignment $\networkInputSet$, we first calculate the values of the declared network variables. 
These values are represented as an \emph{environment}~$\assertEnv : (\nameVar : \tensorTypeVar) \rightarrow \semTensorTheory{\delta}$, i.e.~a function mapping each name $\nameVar$ of the declared input/hidden/output network variable of type $\tensorTypeVar$ to a mathematical tensor of the corresponding type.

Given a list of network declarations $\networkDeclSet$, a set of models $\networkImplementationSet$ and an input assignment $\networkInputSet$, the environment $\assertEnv (\networkDeclSet, \networkImplementationSet, \networkInputSet)$ is calculated as the union of the environments generated by each network declaration:
\begin{equation*}
\assertEnv (\networkDeclSet, \networkImplementationSet, \networkInputSet) = \bigcup_{\networkDeclVar \in \networkDeclSet} \assertEnv (\networkDeclVar, \networkImplementationSet[\field{\networkDeclVar}{name}], \networkInputSet)
\end{equation*}
Concretely, given a network declaration~$\networkDeclVar$, a compatible model~$\networkImplementation$ and input assignment~$\networkInput$, the calculation of the local environment can be defined as follows:
\begin{align*}
\assertEnv (\namedNetworkExpansion{\nameVar'}, \networkImplementation, \networkInputSet) =
\qquad\qquad\qquad\qquad\qquad\qquad
\\ 
\bigcup_{i} \assertEnv_{I}(\inputDeclSet_i) \; \cup \;
\bigcup_{i} \assertEnv_{H}(\hiddenDeclSet_i) \; \cup \; 
\bigcup_{i} \assertEnv_{O}(\outputDeclSet_i, \outputNodesFn(m)_i)
\end{align*}
where
\newcommand{\nameFn}[1]{\text{name}(#1)}
\newcommand{\hiddenNameFn}[1]{\text{hiddenName}(#1)}
{\small
\begin{align*}
    \sem{\networkInput} 
    &= \{\semTensorTheory{\networkInputSet[\field{\inputDeclVar}{name}]} \mid \inputDeclVar \in \inputDeclSet \}
    \\
    \assertEnv_{\inputDeclSet}(\inputExpansion) 
    &= \{ \nameVar \rightarrow \semTensorTheory{\networkInputSet[\nameVar]} \}
    \\
    \assertEnv_{\hiddenDeclSet}(\hiddenExpansion) 
    &= \{ \nameVar \rightarrow \semModelTheory{\networkImplementation}(\sem{\networkInput}, \onnxNameVar) \}
    \\  
    \assertEnv_{\outputDeclSet}(\outputExpansion,\, \onnxNameVar) 
    &= \{ \nameVar \rightarrow \semModelTheory{\networkImplementation}(\sem{\networkInput}, \onnxNameVar) \}
\end{align*}}

\paragraph{Assertions.} Given an environment $\assertEnv$, the semantics of assertions and boolean expressions are defined as: 
\begin{equation*}
    \newcommand{\semComp}[2]{
        \semBool{\mathtt{#1}~a_1~a_2}
        &= \sem{#2}_\networkTheoryVar(\semArith{a_1},\semArith{a_2})
    }
    \begin{array}{llll}
        \semAssert{\mathtt{assert}~b} &= \semBool{b}
        \\
        \semBool{\mathtt{and}~b_1~...~b_n} &=\bigwedge_{i} \;\semBool{b_i}
        \\
        \semBool{\mathtt{or}~b_1~...~b_n} &=\bigvee_{i} \; \semBool{b_i}
        \\
        \semComp{<}{<}
        \\
        \semComp{>}{>}
        \\
        \semComp{<=}{\leq}
        \\
        \semComp{>=}{\geq}
        \\
        \semComp{==}{=}
        \\
        \semComp{!=}{\neq}
    \end{array}
\end{equation*}\unskip
Note that the exact semantics of comparisons is delegated to that of the underlying network theory $\networkTheoryVar$ to ensure the correct behaviour for e.g. floating point comparisons.
The semantics of arithmetic expressions, which are interpreted as zero-dimensional tensors, is defined as follows:
\begin{equation*}
    \begin{array}{llll}
        \semArith{d}
        &= \semTensorTheory{d}
        \\
        \semArith{v[l_1, \ldots, l_k]} 
        &= \assertEnv(v)_{l_1, \ldots, l_k} 
        \\
        \semArith{{\mathtt{-}}~e}
        &= \sem{-}\networkTheoryVarParam(\semArith{e}) 
        \\
        \semArith{\mathtt{+}~e_1~...~e_n} 
        &= \sem{\sum^n_{i=1}}\networkTheoryVarParam~\semArith{e_i}
        \\
        \semArith{\mathtt{*}~a_1~...~a_n} 
        &= \sem{\prod^n_{i=1}}\networkTheoryVarParam~\semArith{e_i}
        \\
        \semArith{\mathtt{-}~a_1~...~a_n} 
        &= \sem{+}\networkTheoryVarParam(\semArith{e_1}, \sem{-}\networkTheoryVarParam(\sem{\sum^n_{i=2}}\networkTheoryVarParam~\semArith{e_i}))
    \end{array}
\end{equation*}\unskip
where $\sem{\prod}\networkTheoryVarParam$ and~$\sem{\sum}\networkTheoryVarParam$ are defined in the obvious way in terms of $\sem{+}\networkTheoryVarParam$ and~$\sem{\times}\networkTheoryVarParam$. As with comparisons, the actual mathematical operations are delegated to the semantics provided by the underlying network theory $\networkTheoryVar$. Constants are interpreted as zero-dimensional tensors and the values of variables are looked up in the environment $\assertEnv$ and then indexed into appropriately.

\paragraph{Queries.}
The semantics of a well-typed query $\queryVar$ is formally defined as a function. The input to the function is a set of models $\networkImplementationSet$ from the underlying theory $\networkTheoryVar$ that are well-typed with respect to $\queryVar$. Its output is a truth value indicating whether there exists an input assignment consisting of a set of tensors~$\networkInputSet$ from the underlying theory that satisfies all the assertions concurrently:
\begin{equation*}
\semQuery{\queryExpansion} = (\networkImplementationSet : \mgrammar{<model>}_\networkTheoryVar^+) \rightarrow
    \exists~(\networkInputSet : \mgrammar{<tensor>}_\networkTheoryVar^+). \; \bigwedge_{\assertionVar \in \assertionSet}~ 
    \semApp{assert}{\assertionVar}^{\assertEnv(\networkDeclSet, \networkImplementationSet,\networkInputSet)} \\
\end{equation*}
Or less formally: given a set of networks does there exist an input assignment under which all the assertions to evaluate to true?
A query is \textit{satisfiable} if such an assignment exists, and \textit{unsatisfiable} otherwise.

\section{Queries over \real}
\label{sec:real-queries}

It is well known that many existing solvers implicitly treat ONNX models as denoting functions over the real numbers, $\mathbb{R}$, and are therefore not sound with respect to the implicit floating-point semantics used by the ONNX standard at runtime~\cite{jia2021exploiting}. 
In the terminology of Section~\ref{sec:network-theory}, such solvers can be understood as interpreting network models using their own semantic function $\semApp{realModel}{\cdot}\networkTheoryVarParam$, rather than the true model interpretation $\semModelAbs\networkTheoryVarParam$ supplied by the underlying network theory~$\networkTheoryVar$.
Although this discrepancy is problematic from a soundness perspective, it would be equally undesirable to simply declare these solvers as non-compliant with VNN-LIB~2.0.
Instead, we argue that potentially unsound real-valued analyses should be supported by the standard, provided that the query author explicitly authorises it.

\begin{figure}[tp]
    \begin{minipage}[c]{0.62\textwidth}
        \begin{lstlisting}[style=lbnf]
(declare-network myNetwork
  (declare-input  X real [1,10])
  (declare-output Y real [1,2]))

(assert (>= X[0,2] 0.0))
(assert (<= X[0,2] 1.0))
(assert (<= Y[0,1] 0.5))\end{lstlisting}
    \end{minipage}%
    \begin{minipage}[c]{0.35\textwidth}
        \centering
        \includegraphics[height=4cm, alt={A visualisation of a neural network model with two nodes.}]{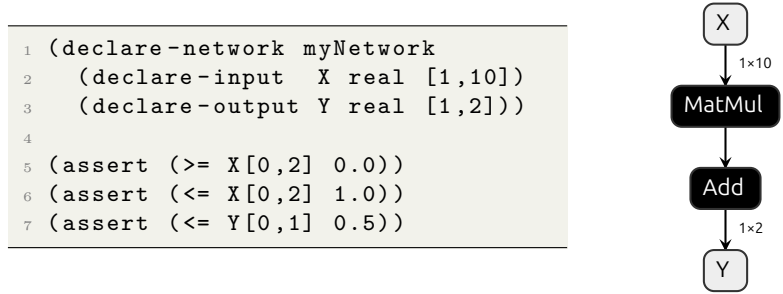}
    \end{minipage}
    \caption{An example of a real-valued \vnnlib{} query. This explicitly gives the solver permission to unsoundly interpret the models from the network theory~$\networkTheoryVar$ as functions over \real, even though in actuality the models will be executed using floating-point types.}
    \label{fig:real-simple-query}
\end{figure}

To this end, the \vnnlib{} standard introduces the notion of \emph{real-valued queries}, illustrated in Figure~\ref{fig:real-simple-query}.
Unlike standard queries which use the element types supplied by the network theory~$\networkTheoryVar$, real-valued queries are required to exclusively use the distinguished type \inlinevnn{real}.
The use of this type indicates explicit permission for the solver to interpret ONNX networks as real-valued functions
and acknowledges that the result may be unsound due to numerical imprecision. 
This approach enables existing solvers based on real-valued reasoning to participate in a controlled manner, while providing a clear migration path towards fully sound floating-point verification.

\begin{figure}[t]      
    \newcommand{\realParam}{\hspace{-0.25em}$^\real$}
	\begin{subfigure}{\textwidth}
        \centering
		\input{diagrams/real-network-theory-syntax-types.tex}
		\caption{Grammar for real network model types.}
		\label{fig:real-theory-type-syntax}
	\end{subfigure}
	\vspace{0.5em}
	\\
    \begin{subfigure}{\textwidth}
	    \centering
    	\input{diagrams/real-network-theory-syntax-expr.tex}
		\caption{Grammar and functions for real network model expressions.}
		\label{fig:real-theory-expr-syntax}
	\end{subfigure}
	\vspace{0.3em}
    \\
    \begin{subfigure}{\textwidth}
        \input{diagrams/real-network-theory-grammar.tex}
		\caption{Type-system for real network model expressions.}
		\label{fig:real-theory-grammar}
	\end{subfigure}
	\vspace{0.2em}
    \\
    \begin{subfigure}{\textwidth}
        \centering
		\input{diagrams/real-network-theory-semantics.tex}
		\vspace{-1.7em}
		\caption{Semantics for real network model expressions.}
		\label{fig:real-theory-semantics}
	\end{subfigure}
    \caption{Constructing a real-valued network theory $\networkTheoryVar_\real$ from an abstract network theory~$\networkTheoryVar$ and a function $\semApp{realModel}{\cdot}\networkTheoryVarParam$ that returns the real-valued semantics of a network model. Semantics for $\sem{\leq}$, ..., $\sem{\times}$ (not shown) simply delegate to standard operations over $\real$.}
    \label{fig:real-network-theory}
\end{figure}

The remaining challenge is to give a precise mathematical account of real-valued queries.
The notion of a network theory provides a modular mechanism for defining both the syntax and semantics of such queries.
Given a network theory~$\networkTheoryVar$ and a function $\semApp{realModel}{\cdot}\networkTheoryVarParam$ interpreting models in $\networkTheoryVar$ as real-valued functions, we can define a new network theory $\networkTheoryVar_\real$ as shown in \autoref{fig:real-network-theory}. In particular:
\begin{itemize}
\item \textbf{Element types} in $\networkTheoryVar_\real$, consist of a single type, namely \texttt{real}, that has the expected typing rules and is interpreted in the semantics as $\real$.
\item \textbf{Tensors} in $\networkTheoryVar_\real$ are syntactically just mathematical tensors over the real numbers, again with the expected typing rules and semantics.
\item \textbf{Models} - models in $\networkTheoryVar_\real$ are models from the original theory $\networkTheoryVar$. They are well-typed if the shapes of the input and output tensors match, i.e. the actual element types from $\networkTheoryVar$ used by the model are irrelevant. The semantics of the model is provided by $\semApp{realModel}{\cdot}\networkTheoryVarParam$.
\end{itemize}
We can then use this new theory $\networkTheoryVar_\real$, to immediately obtain both the syntax and semantics for real-valued queries simply by reparametrising the existing syntax and semantics defined in the previous sections. 
This provides yet another powerful demonstration of the utility of the network theory abstraction.
\section{Formalisation in Agda}
\label{sec:agda}

To provide assurance in the consistency of the standard itself, we have mechanised Version~2.0 of VNN-LIB in the interactive theorem prover Agda~\cite{norell2009dependently}. 
This formalisation enforces internal consistency between the syntax and semantics of the language and rules out ambiguities that are difficult to detect in the informal written version of the specification presented in the previous sections. 
Moreover, the formalisation provides a potential basis for machine-checked proofs about tools that consume or produce VNN-LIB, including compilers, solvers and query optimisers. 
The development has no dependencies other than the Agda standard library~\cite{daggitt2025agda} and is provided as an artifact accompanying this paper.
\section{Additional Features}

Several features of \vnnlib{}~2.0 fall outside the scope of this paper.
\begin{enumerate}
    \item Due to space constraints, we have omitted a description of the logics and theories that identify fragments of the language supported by different classes of solvers, analogous to the logics and theories defined in SMT-LIB.
    \item \vnnlib{}~2.0 specifies a standardised command-line interface that enables higher-level tools to invoke solvers and to query their capabilities. As example of the latter, a tool can indicate via the command line that it only supports queries over single networks and therefore will not accept queries with multiple network declarations. 
    This allows solver developers to  migrate to \vnnlib{}~2.0 without being forced to add support the new advanced query features such as multiple networks, hidden nodes or non-linear arithmetic.
    \item In addition to the Agda  formalisation described in Section~\ref{sec:agda}, the new version of the standard is accompanied by a high-performance C++ library implementing a parser and type checker compliant with Sections~\ref{sec:syntax} and~\ref{sec:typing}, along with Python and Julia bindings. 
    We are aware of efforts to build official Rocq and Ocaml bindings as well.
    These libraries will reduce the barrier for entry for both new tools and existing tools seeking to support  \vnnlib{}~2.0.
\end{enumerate}
For a fuller description of these components, readers should consult the official VNN-LIB standard document and the VNN-LIB website. 

\section{Community Response}

The neural network verification community was consulted extensively during the development of VNN-LIB~2.0, and the feedback received has been overwhelmingly positive. 
The teams behind several leading solvers have already committed to supporting the new standard, and VNN-LIB~2.0 benchmarks will be used in VNN-COMP~2026. 
We expect the standard to continue evolving, and we have already received proposals for further extensions, such as support for network derivatives and a standardised format for proof certificates.

\section{Conclusion}

In summary, this paper has made two complementary contributions to the theory and practice of neural network verification.
First, we have substantially extended the expressiveness of neural network verification queries, enabling the community to specify and verify richer properties beyond traditional robustness in a standardised manner.
This will make tools that support such specifications easier to use and will also allow their performance to be evaluated and benchmarked (e.g. in VNN-COMP), thereby driving further progress in algorithm design.

Second, we have provided rigorous formal semantics for neural network verification queries. 
While the semantics may appear theoretical, they have many concrete practical consequences. 
They ensure that higher-level tools and solvers can trust that queries are interpreted unambiguously, provide the technical language needed to identify and resolve subtle issues (as illustrated in Section~\ref{sec:real-queries} for real-valued queries), and establish a rigorous contract with solvers for the expected numeric behaviour of a query.
Moreover, the semantics form a crucial step towards formally verifying existing neural network verification tools, as well as developing new tools such as formally verified proof checkers for neural network verification certificates~\cite{elboher2025abstraction}.

\paragraph{Acknowledgements}
Version 2.0 of the standard was developed with the
help of many others in the neural network verification community. Particular
thanks must given to the following people, who provided many helpful suggestions, constructive criticism and encouragement: Taylor Johnson, Samuel Teuber,
Wen Kokke, Julien Girard, Guilhem Ardouin, Augustin Lemesle, Michele Alberti,
Julien Lehmann, Thomas Flinkow, Edoardo Manino, Guy Amir, Omri Isac, Guy
Katz, Idan Refaeli, David Shriver and Christopher Brix.

\paragraph{Data availability}
As well as being actively under development on GitHub, the frozen version of the Agda formalisation designed to accompany this paper is available as a Docker image on Zenodo with DOI \href{https://doi.org/10.5281/zenodo.19858445}{10.5281/zenodo.19858445}.

\paragraph{Contributions}
\textbf{Ann Roy}: syntax and semantics, Agda formalisation, document preparation. \textbf{Allen Anthony}: syntax, C++ and Python libraries, document preparation. \textbf{Andrea Gimelli}: syntax, artefact preparation and document preparation.
\textbf{Matthew Daggitt}: conceptualisation, syntax and semantics, Agda formalisation, document preparation.

\paragraph{Disclosure of interests}
The authors have no competing interests to declare that are
relevant to the content of this article.

\bibliographystyle{splncs04}
\bibliography{bibliography}
\end{document}